\pdfoutput=1
\documentclass[11pt, utf8]{article}
\usepackage{pgfplots}
\pgfplotsset{compat=1.17}
\usepackage{pgf-pie}
\usepackage[bottom]{footmisc}
\usepackage{footnote}
\usepackage{pdfpages}
\DeclareUnicodeCharacter{0307}{\d}

\usepackage[final]{acl}
\usepackage{amsmath}
\usepackage{amsfonts}
\usepackage{booktabs}
\usepackage{float}
\usepackage{times}
\usepackage{latexsym}
\usepackage[T1]{fontenc}
\usepackage[utf8]{inputenc}
\usepackage{microtype}
\usepackage{inconsolata}
\usepackage{graphicx}
\title{DiscoGraMS: Enhancing Movie Screen-Play Summarization using Movie Character-Aware Discourse Graph}
\author{
Maitreya Prafulla Chitale$^{1}$, Uday Bindal$^{1}$, Rajakrishnan Rajkumar$^1$, Rahul Mishra$^1$ \\
\\
$^1$IIIT Hyderabad \\
\texttt{\{maitreya.chitale, uday.bindal\}@research.iiit.ac.in} \\
\texttt{\{raja, rahul.mishra\}@iiit.ac.in}
}

\begin{document}
\maketitle
\begin{abstract}
    Summarizing movie screenplays presents a unique set of challenges compared to standard document summarization. Screenplays are not only lengthy, but also feature a complex interplay of characters, dialogues, and scenes, with numerous direct and subtle relationships and contextual nuances that are difficult for machine learning models to accurately capture and comprehend. Recent attempts at screenplay summarization focus on fine-tuning transformer-based pre-trained models, but these models often fall short in capturing long-term dependencies and latent relationships, and frequently encounter the \textit{"lost in the middle"} issue. To address these challenges, we introduce \textbf{DiscoGraMS}, a novel resource that represents movie scripts as a movie character-aware discourse graph \textbf{(CaD Graph)}. This approach is well-suited for various downstream tasks, such as summarization, question-answering, and salience detection. The model aims to preserve all salient information, offering a more comprehensive and faithful representation of the screenplay's content. We further explore a baseline method that combines the CaD Graph with the corresponding movie script through a late fusion of graph and text modalities, and we present very initial promising results.
    We have made our code\footnote{https://github.com/Maitreya152/DiscoGraMS} and dataset\footnote{https://huggingface.co/datasets/Maitreya152/CaD\_Graphs} publicly available.
\end{abstract}

\section{Introduction}
\label{sect:intro}
Text summarization has been extensively studied within the NLP community \cite{nallapati-etal-2016-abstractive, Nallapati_summarunner, zheng-lapata-2019-sentence, urlana-etal-2024-controllable}. Recently, large language models (LLMs) have demonstrated human-level performance in this area \cite{liu-etal-2023-revisiting,zhang-etal-2024-benchmarking}. However, summarizing long documents remains a challenge for even the most advanced LLMs, as their effectiveness can be influenced by the location of salient information within the text \cite{liu-etal-2024-lost}. For language models to effectively utilize information within very long input documents, their performance should exhibit minimal sensitivity to the positional placement of relevant information within the input \cite{liu-etal-2024-lost}. Movie script or screenplay summarization \cite{papalampidi-etal-2020-screenplay,saxena-keller-2024-moviesum} is a relatively hard task compared to standard document summarization due multitude of reasons. Movie scripts are typically very long documents characterized by intricate narratives, numerous subplots, and substantial dialogue, which pose significant challenges for summarizing the content without losing the core elements of the story. Many of the movie scripts have non-linear flow of events such as flashbacks, flash-forwards, and parallel plot lines, making the summary to retain the coherence and original flow. 

\begin{figure}[t]
    \centering
    \includegraphics[width=0.93\linewidth]{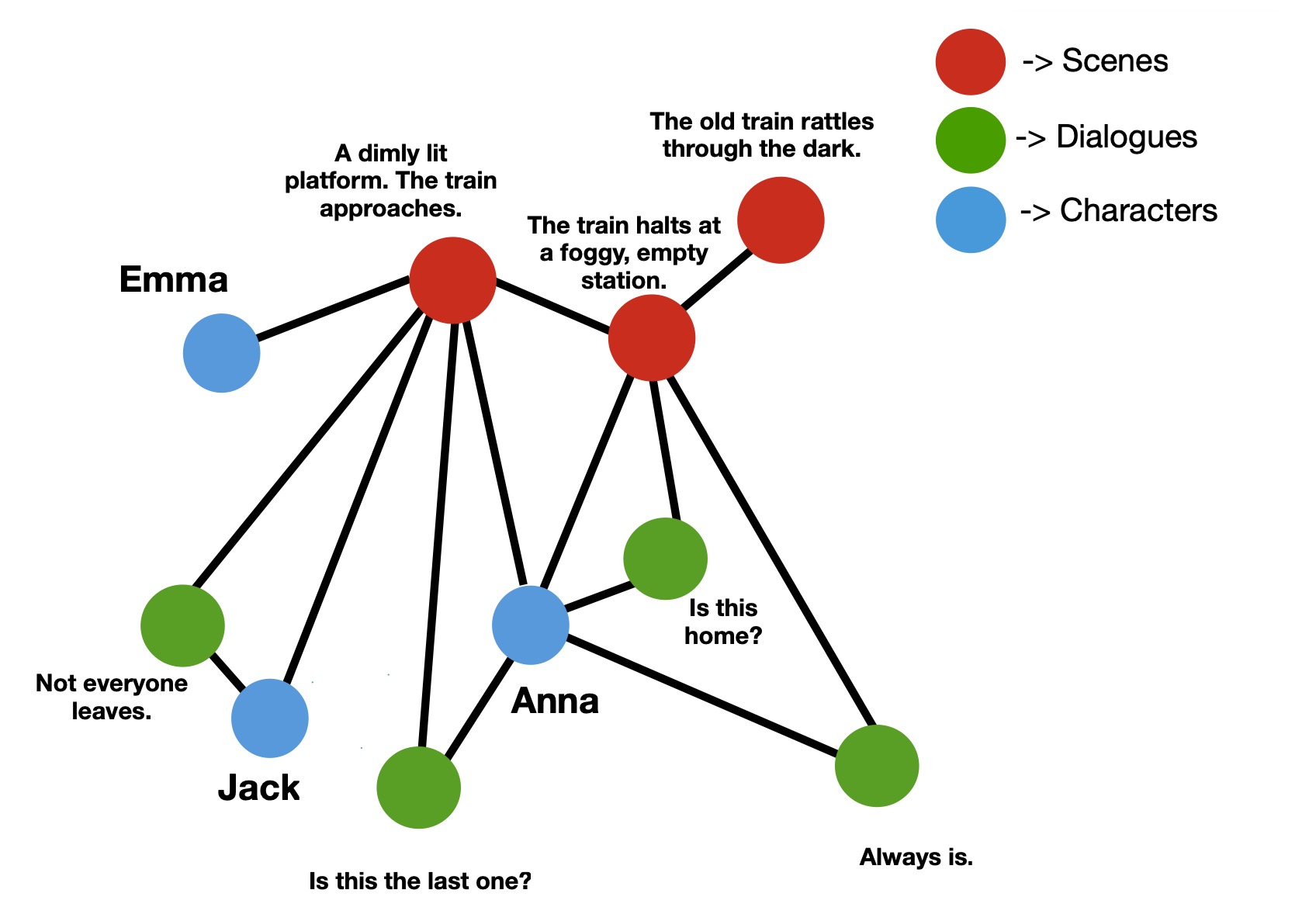}
    \caption{Example of a graph constructed from a movie script.}
    \label{fig:example}
    \vspace{-0.72cm}
\end{figure}

To address this, we present DiscoGraMS, an innovative resource that represents movie scripts as a character-aware discourse graph (CaD Graph). This graph captures the core essence of the movie plot by modeling latent relationships among key elements, including characters, the scenes they participate in, and the dialogues they deliver, thereby highlighting all possible semantically important aspects of the narrative. The CaD Graph captures intricate nuances and the interplay between characters and scene sequences, effectively addressing challenges like flashbacks and sudden plot twists that are difficult to capture using only textual content.    
The main contributions of this work are as follows: 1) We introduce, for the first time to our knowledge, a movie character-aware discourse graph (\textbf{CaD Graph}) specifically designed for movie script summarization.  
2) We propose a \textbf{late modality fusion model} that combines both CaD Graphs and textual content for improved movie script summarization.  
3) We perform an \textbf{ablation study} to demonstrate the effectiveness of CaD Graphs in enhancing summarization.  
\\
\section{Related Work}\label{sect:related-work}
Since the origin of modern graph theory in 1736 with Euler's proof the {\it Seven Bridges of Königsberg} problem {\it i.e.} traversing a city crossing 7 bridges exactly once~\citep{harary1960}, graph representations have been used to model data in diverse fields like chemistry, biology and computer science. Linguistic data has also been represented as graph structures like dependency representations~\citep{Tesniere:1959} and successfully deployed in NLP applications. The idea of representing entire texts as graphs was proposed in seminal work by~\citet{mihalcea-tarau-2004-textrank}. They created graphs comprising of nodes which keywords connected to other words located within a window of 2 to 10 words. This approach was extremely effective for the  task of extractive summarization. More recently,~\citet{wang-etal-2022-multi} show the efficacy of this technique for abstractive summarization of scientific articles. Here, entities in the text served as nodes (with co-referential entity clusters represented as a single node) connected to one another via labelled edges depicting relationships (like hyponymy) between nodes. \cite{kounelis2021movie} proposed a movie recommendation system using character graph embeddings to model relationships for movie similarity while \cite{papalampidi2021movie} propose a model for summarizing movie videos by constructing a sparse graph using only the turning point scenes from videos. In contrast, our CaD Graph method integrates scene, dialogue, and character interactions and focuses on summarizing movie text scripts, which presents a distinct set of challenges due to the long-form nature of screenplay texts. Prior work has explored character-based graphs in narratives. \cite{agarwal2013sinnet} introduced SINNET, a system for extracting social interaction networks from text. \cite{srivastava2016inferring} focused on inferring interpersonal relationships in narrative summaries, while \cite{elson-etal-2010-extracting} developed methods for extracting social networks from literary fiction. \cite{zhao2020bridging} propose DualEnc to bridge the structural gap in data-to-text generation by integrating graph and sequential representations. Our work builds on these approaches by constructing a CaD Graph to enhance screenplay summarization. There have been no significant efforts to employ graphs for movie script summarization. Only recently, \cite{saxena-keller-2024-moviesum} adapted TextRank \cite{zheng-lapata-2019-sentence}, a sentence centrality-based graph approach, for movie scripts. However, this approach was outperformed by the simpler Longformer Encoder-Decoder (LED) model \cite{beltagy2020longformerlongdocumenttransformer} by large margin.

\section{Dataset}
We use the MovieSum \cite{saxena-keller-2024-moviesum} dataset, a comprehensive resource for movie summarization, containing 2,200 movie screenplays along with metadata and plot summaries, including movies up to 2023. The plot summaries are sourced from IMDb and Wikipedia, ensuring a diverse range of writing styles and perspectives. The summaries were generated through a combination of automatic extraction and manual curation by trained annotators. The scripts are in XML format, preserving key elements such as scene descriptions, dialogues, and character names for efficient analysis. The dataset is split into training (1,800 movies), validation (200 movies), and test (200 movies) sets, with average screenplay lengths of 29k words and summaries of 717 words. The summaries, sourced from IMDb and Wikipedia, blend automatic extraction and manual curation. Analysis reveals a high level of abstractiveness in the summaries, indicated by novel 3-grams and 4-grams not found in the original scripts.

\begin{table}[h]
\centering
\scalebox{0.72}{
\begin{tabular}{cccc}
\hline
\multicolumn{4}{c}{\textbf{\%  Novel n-grams in Summary}} \\
\hline
1-grams & 2-grams & 3-grams & 4-grams \\
\hline
31.69 & 68.88 & 93.12 & 98.6 \\
\hline
\end{tabular}
}
\caption{Percentage of novel n-grams in summary. \cite{saxena-keller-2024-moviesum}}
\label{tab:novel-ngrams}
\vspace{-0.5em}
\end{table}

\begin{figure*}
    \hspace*{1cm}
    \includegraphics[width=0.93\linewidth]{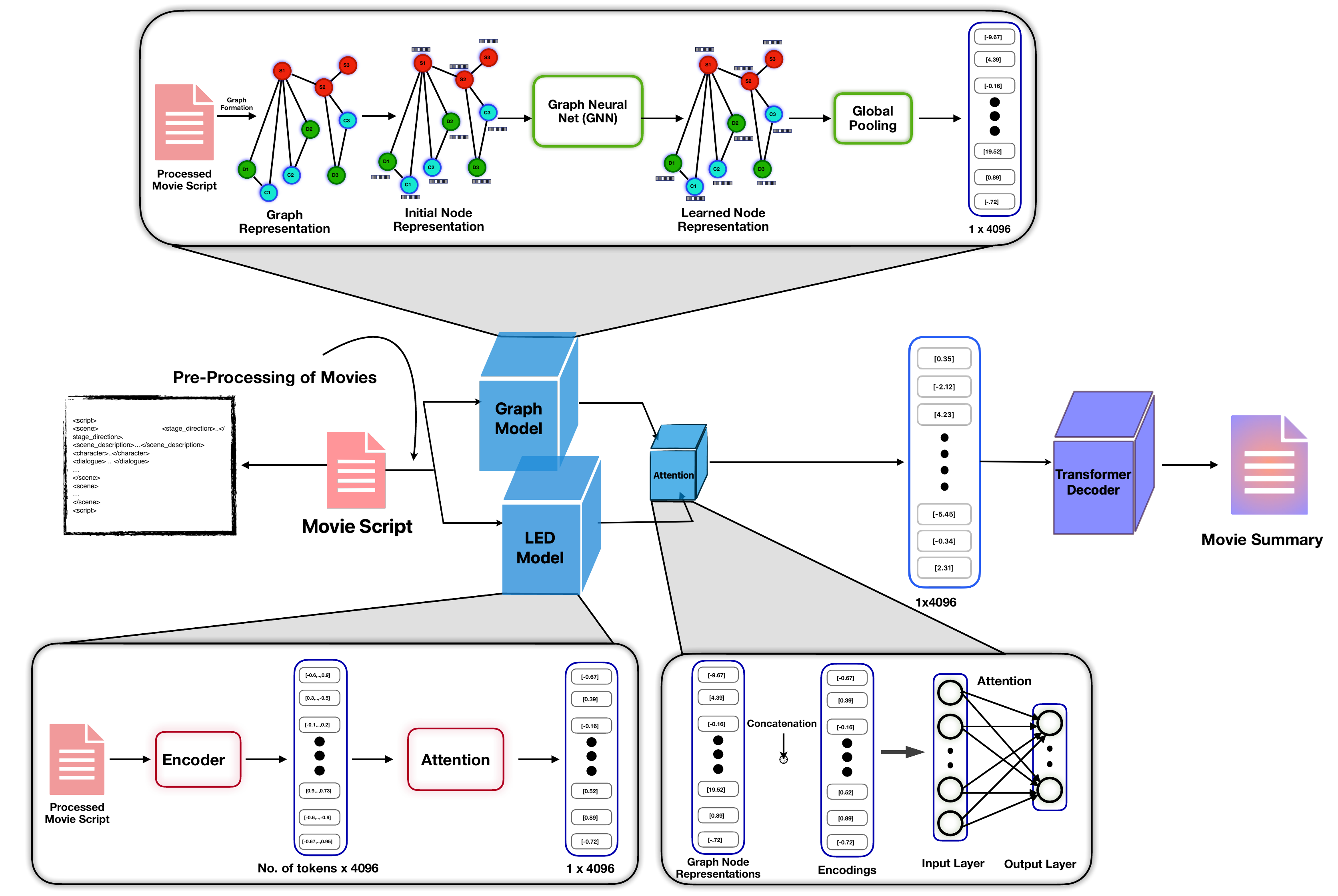}
    \caption{Architecture Diagram for the proposed model LGAT.}
    \label{fig:baselineArch}
\end{figure*}
\section{Methodology}
\label{sec:methodology}
In this section, we describe the process of constructing the character-aware discourse graph (CaD Graph) from movie scripts. We then present a baseline method that leverages both the CaD Graph and the textual content of the scripts, using a late modality fusion approach to generate movie script summaries.
\subsection{Graph Construction and Encoding:}
The first step involves constructing a graph representation of the movie script. In this representation, nodes are created for key elements, which are scenes, characters, and their dialogues.\\
The constructed graph can be described as a heterogeneous graph \( G = (V, E) \), where \( V \) is the set of nodes, and \( E \) is the set of edges. There are three types of nodes, scenes (\( V_s \)), dialogues (\( V_d \)), and characters (\( V_c \)). The edges represent different relationships, \( E_{ss} \subseteq V_s \times V_s \): Edges between consecutive scenes, \( E_{sd} \subseteq V_s \times V_d \): Edges between scenes and dialogues occurring in those scenes, \( E_{sc} \subseteq V_s \times V_c \): Edges between scenes and characters appearing in those scenes, \( E_{cd} \subseteq V_c \times V_d \): Edges between characters \indent \indent and dialogues spoken by those characters. Formally, the graph construction is written as follows:
$
G = \left( V_s \cup V_d \cup V_c, E_{ss} \cup E_{sd} \cup E_{sc} \cup E_{cd} \right)$ \\
\textbf{Scene Nodes}:
    $
    V_s = \{ s_i \mid s_i \text{ is a scene} \}
    $
    Each scene node \( s_i \) has an associated embedding \( \mathbf{e}(s_i) \) representing the scene description text, derived from the sentence embedding model (SE) \cite{reimers-2019-sentence-bert}:
    $
    \mathbf{e}(s_i) = \text{SE}(\text{Scene Description}(s_i))
   $
    The scenes list is ordered according to the order in which the scenes occur in the movie.\\
\textbf{Dialogue Nodes}:
    $
    V_d = \{ d_j \mid d_j \text{ is a dialogue} \}
    $
    Each dialogue node \( d_j \) has an associated embedding \( \mathbf{e}(d_j) \), representing the dialogue text:
    $
    \mathbf{e}(d_j) = \text{SE}(\text{Dialogue Text}(d_j))
    $\\
\textbf{Character Nodes}:
    $
    V_c = \{ c_k \mid c_k \text{ is a character} \}
    $
    The characters are initialised with zero embedding whose dimension matches with the embedding dimension of the sentence encoder.\\
\textbf{Edges}: The edges between the scenes and other entities are defined as follows: the scene-to-scene edges are given by $E_{ss} = ( (s_i, s_{i+1}) \mid s_i, s_{i+1} \in V_s )$ the scene-to-dialogue edges are defined as $E_{sd} = \left( (s_i, d_j) \mid d_j \in V_d, \, s_i \in V_s, \, d_j \textit{ occurs in scene } s_i \right)$ the scene-to-character edges are defined as $E_{sc} = \left( (s_i, c_k) \mid c_k \in V_c, \, s_i \in V_s, \, c_k \textit{ occurs in scene } s_i \right)$ and finally, the character-to-dialogue edges are given by $E_{cd} = \left( (c_k, d_j) \mid c_k \in V_c, \, d_j \in V_d, \, d_j \textit{ is spoken by } c_k \right)$.
\\\\
A movie's CaD Graph consists of intricate connections that represent the three-way relationships between scenes, characters, and dialogues, as illustrated in Figure \ref{fig:example}. Adding sequential links between scenes helps the model capture the movie's overall flow. The connections from scenes to characters and dialogues to characters enable the model to differentiate between characters and understand their roles. We hypothesize that this structure also helps the model infer a character's significance within the movie, making our graphs, DiscoGraMS, character-aware.

\subsection{The Proposed Model LGAT}

We propose a novel late fusion-based model, LGAT, which integrates the CaD Graph and the textual content of movie scripts through a Graph Neural Network (GNN) using graph attention with convolutions and a Longformer Encoder-Decoder (LED) \cite{beltagy2020longformerlongdocumenttransformer} text encoder, as illustrated in Fig \ref{fig:baselineArch}. This combination generates the script's encoding, followed by a decoder that produces the summary. A detailed explanation of the model's internals is provided in Appendix Sec \ref{appen: baseline-details} due to space limitations.

\section{Results}
We select the models \textbf{LongT5} \cite{guo-etal-2022-longt5}, \textbf{PEGASUS-X} \cite{phang-etal-2023-investigating}, and the Longformer Encoder-Decoder
(\textbf{LED}) model \cite{beltagy2020longformerlongdocumenttransformer}, (See Table \ref{tab:baselines}) as the baselines (inspiration for baselines are drawm from \cite{saxena-keller-2024-moviesum}) to compare with our proposed model.

\begin{table}[h]
\centering
\scalebox{0.63}{
\begin{tabular}{lccccccc}
\toprule
\textbf{Model}                & \textbf{R-1 \( \uparrow \)}   & \textbf{R-2 \( \uparrow \)}   & \textbf{R-L \( \uparrow \)}   & \textbf{BS$_p$ \( \uparrow \)} & \textbf{BS$_r$ \( \uparrow \)} & \textbf{BS$_{f1}$ \( \uparrow \)} \\ \midrule
\multicolumn{7}{c}{\textbf{Baseline Models}} \\ \midrule
Pegasus-X 16K        & {42.42} & 8.16  & {40.63} & {58.81}  & {56.06}  & 54.36   \\
LongT5 16K           & 41.49 & {8.39}  & 39.78 & 56.09  & 55.60  & {55.68}   \\
Longformer (LED) 16K & \underline{44.85} & \underline{9.83}  & \textbf{43.12} & \underline{59.11}  & \underline{58.43}  & \underline{58.73}   \\ \midrule
\multicolumn{7}{c}{\textbf{Proposed Model}} \\ \midrule
\textbf{LGAT (Ours)}       & \textbf{49.25} & \textbf{13.12}  & \underline{34.61} & \textbf{80.68}  & \textbf{82.36}  & \textbf{81.51}   \\ \bottomrule
\end{tabular}
}
\caption{Comparison of Baseline Models and Proposed LGAT Model on the test set. The results of the baselines are referred to from \cite{saxena-keller-2024-moviesum}. Best scores are \textbf{bold}. Second Best scores are \underline{underlined}. \( \uparrow \) Indicates higher values are better.}
\label{tab:baselines}
\end{table}
The proposed model has the following configuration: LongFormer Encoder (LE) 4K + GATConv (\textbf{LGAT}), Where \textbf{LE} \cite{beltagy2020longformerlongdocumenttransformer}, is the longformer encoder. We use 4K context window for LED only compared to 16K used in MovieSum \cite{saxena-keller-2024-moviesum} due to limited compute resources (Appendix \ref{appen: implementation-details}) availability, The results for this experiment can be obtained in Table \ref{tab:baselines}.

As presented in Table \ref{tab:baselines}, our proposed model, LGAT, significantly outperforms all baseline models on both ROUGE and BERT score metrics. This improvement can be attributed to the cues and patterns provided by the CaD Graph, which capture the overall essence of the movie plot. However, we observe that for the ROUGE-L metric, LGAT does not surpass the LED baseline, likely due to the smaller context window used in our encoder (4K vs. 16K). 

\subsection{Ablation Studies}
The \textbf{LE} architecture, along with \textbf{GATConv}, has proven to be suited for processing long sequences. Following this, we run ablation studies on \textbf{LGAT} to prove the effectiveness of our proposed architecture of combining \textbf{GATConv} and \textbf{LE}. Specifically, we train both the encoders decoupled and test them on the test set. We compare the results against the full model (LGAT) to prove the effectiveness of the individual parts of the architecture, and hence show how they individually contribute towards the final result. To further strengthen our hypothesis regarding the importance of incorporating character information in the graph, we perform an additional ablation study. Specifically, we remove all character-related nodes and edges from the graph and evaluate the performance of the model in this modified setup. This ablation isolates the impact of character awareness in the graph structure and provides insight into the contribution of character-related information to the model’s effectiveness. The results for this ablation study can be found in Table \ref{tab:ablation}. We observe that GNN-based CAD graph encoding is very useful andcontributing more than LED-based textual encoder. Moreover, it is proved that character-awareness has a positive impact towards the performance of the model.
\begin{table}[ht]
\centering
\scalebox{0.63}{
\begin{tabular}{lccccc}
\toprule
\textbf{Model \( \uparrow \)}                & \textbf{R-1}   & \textbf{R-2}   & \textbf{R-L}   &  \textbf{BS$_{f1}$} \\ \midrule
LE       & 16.16 & 1.63  & 13.20 & 71.95   \\
GATConv           & {43.60} & {8.91}  & {28.70} &  {79.07}   \\ 
LGAT (Without Characters) & \underline{45.99} & \underline{10.78} & \underline{30.61} & \underline{80.31} \\
LGAT (Full)     & \textbf{49.25} & \textbf{13.12}  & \textbf{34.61} & \textbf{81.51}   \\\bottomrule
\end{tabular}
}
\caption{Results of Ablation Studies in comparison to our full model. Best Scores are \textbf{bold}. Second Best Scores are \underline{underlined}. \( \uparrow \) Indicates Higher The Better for all scores.}
\label{tab:ablation}
\vspace{-1em}
\end{table}

\section{Discussion}\label{sect:disc}

\begin{figure}
    \centering
    \includegraphics[width=1\linewidth]{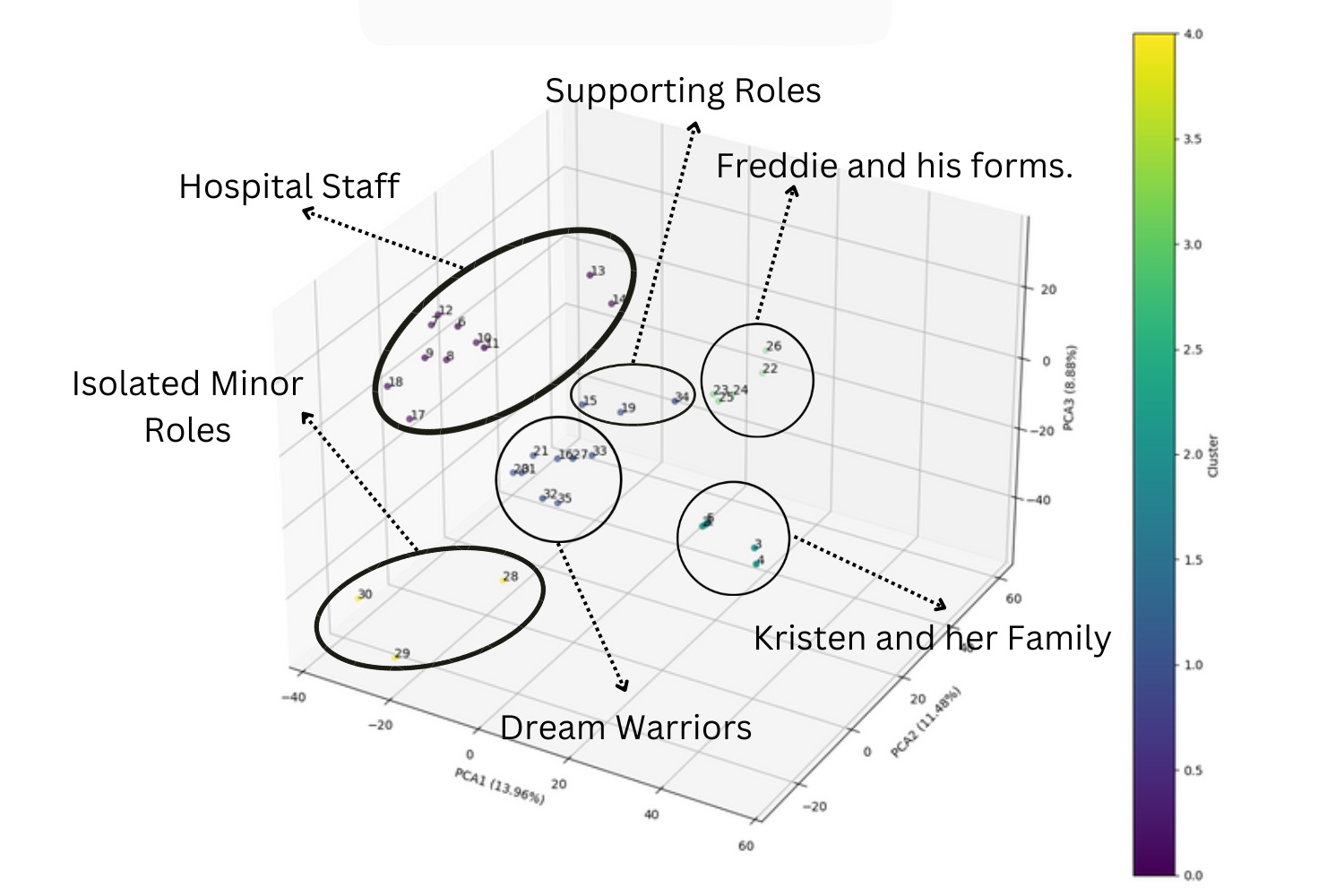}
    \caption{Character Embeddings from the Movie: A Nightmare on Elm Street 3: Dream Warriors. Sections are annotated with the class of characters that is a majority within them.}
    \label{fig:character embeddings}
\end{figure}

Our experiments on abstractive summarization of movie screenplays ({\it i.e.}, the process of generating a plot summary given a screenplay) show that representing screenplays as graphs consisting of scenes, dialogues, and characters holds a lot of promise for movie summarization.
To show how our character-aware graphs capture the roles of different characters in the graph and represent them, we plot the extracted node embeddings from our model in Figure \ref{fig:character embeddings}. First, the node embeddings of all the nodes in the graph are extracted by passing the required movie graph over the GNN part of the final trained model. Next, the acquired node embeddings are filtered so that they only contain the node embeddings of the movie characters, and other node embeddings such as those of scenes and dialogues are discarded. Once the character node embeddings are extracted, they are analyzed using Principal Component Analysis (PCA) to reduce their dimensionality while preserving essential variance. We employ PCA to project the high-dimensional embeddings into a three-dimensional space, allowing for better visualization and interpretability. The transformed embeddings are then clustered using the K-Means algorithm, which groups characters into distinct clusters based on their learned representations.

To further illustrate the relationships and roles of different characters, we visualize the clusters in a three-dimensional scatter plot, where each point represents a character, and the color corresponds to the assigned cluster through K-Means clustering. This visualization enables us to observe meaningful patterns in the character representations. Characters who frequently interact or share similar narrative functions often appear closer together, whereas those with distinct roles are more clearly separated. The clustering also helps to reveal latent groupings, such as protagonists, antagonists, and supporting characters, as also depicted in Figure \ref{fig:character embeddings}. This demonstrates how our approach successfully captures narrative structures through graph-based representation learning.

The effectiveness of our method in clustering and analyzing character embeddings suggests that our GNN-based approach learns informative representations that reflect underlying narrative and character dynamics.

\section{Conclusion}\label{sect:concl}
Our approach outperforms quantitative results (except R-L) reported in prior work on movie summarization using the same dataset~\cite{saxena-keller-2024-moviesum}. We attribute the better performance of our system to the presence of richer graphs, and encoding schemes. Specifically, we attribute the phenomenal improvement in BERT Score to the introduction of an attention layer to combine the encodings of the chunks as discussed in Section \ref{sec:ms_enc} and the novel \textbf{CaD Graph} which enables the model to easily retain salient information which is validated by the high BERT Scores. We suspect that the low scores obtained in R-L are mainly due to the lower context size model (LED 4K) due to a restriction on the available compute resources. The model's (LED) low performance in isolation validates our believes. Our results indicate that knowledge-based representations of the text and plot structure help deep learning algorithms.

We expect our approach to have implications for other NLP problems like Question-Answering, Genre Identification, and Saliency Detection. ~\cite{xu-etal-2024-fine} propose a system to represent narrative text consisting of passages as nodes connected by edges encoding cognitive relations between them. In addition to mainstream engineering applications, our graph representations can be deployed in scientific studies of inferencing processes in narrative comprehension by humans. 


\section*{Limitations}
Our graphs are devoid of co-reference resolution strategies which can take insights from the referred characters and add crucial information about the movie plot.
In addition to this, we were inhibited by our lack of compute resources, due to which we were not able to load the LED 16K model to encode movie scripts. This lack of compute resources also limited our choice of \textit{architecture\_dim} which is capped at 4K.  This constraint potentially impacts the Rouge-L scores, resulting in lower performance. 
We were unable to conduct graph ablations (specifically, the removal of character and dialogue nodes) to evaluate their individual contributions to the model's performance. 
In future work, we plan to address these.


\section*{Ethics Statement}
\textbf{Dataset:} Even though metadata and summaries of each movie are sourced from public domains (wikipedia, imdb), privacy and copyright considerations have been respected. Care has been taken so no sensitive or personally identifiable information is included. The movie scripts may reflect bias to particular genres or cultural context which may affect the behavior of the model. \\
\\
\textbf{Language Models:} The paper includes the usage of pre-trained language models for the task of generating embeddings (section \ref{sec:methodology}). These models are susceptible to biases inherent in their training data . As a result, any summaries produced from our model should be subject to manual review before being released.

\bibliography{main}
\appendix

\section{Details of the Proposed Model}
\label{appen: baseline-details}
The constructed CAD graph is subsequently encoded using a Graph Attention Network (GATConv in PyTorch Geometric \footnote{\url{https://pytorch-geometric.readthedocs.io/en/latest/generated/torch_geometric.nn.conv.GATConv.html}}
) \cite{veličković2018graphattentionnetworks}. This encoding process helps in capturing complex relationships and contextual information inherent in the graph structure. The resulting graph embeddings provide a rich representation of not only the interconnections among scenes, characters, and dialogues, but also the information contained within the scenes, and dialogues.\\
The choice of a GATConv was made by keeping in mind that not all scenes, dialogues, or characters, are equally important and should be included in the summary. Thus, a convolution method which attends differently to different nodes was an ideal choice for this.
\subsection{Movie Script Encoding:}
\label{sec:ms_enc}
We employ the longformer encoder to generate embeddings for the textual content of the movie script.\\
First, the entire script is divided into chunks, with each chunk sized according to the maximum input length the encoder can process.\\
Each chunk is then passed through the encoder, producing an encoding of shape \textit{\({[ \text{chunk\_size}, \text{ max\_tokens}, \text{ encoding\_dim} ]}\)}, where \textit{encoding\_dim} refers to the dimensionality of the encoder.\\
Finally, these embeddings are transformed into a single embedding of shape \textit{\({[ 1, \text{ architecture\_dim} ]}\)} via a \textbf{multi-headed self-attention layer} \cite{vaswani2023attentionneed}. Here, \textit{architecture\_dim} is a hyper-parameter, as described in Appendix~\ref{appen: implementation-details}, it also represents the final embedding dimension for the movie.\\
We hypothesize that by applying multi-headed self-attention, the resulting compressed embedding will effectively capture the most relevant parts of the movie for the purpose of summarization.
\subsection{Encoding Integration:}
After obtaining the encodings from both the Graph Encoder Model and the Text Encoder Model, we perform a \textbf{concatenation} of these representations and then pass it through another \textbf{multi-headed self-attention layer}. This integration facilitates an effective combination of features and relations derived from the graph as well as the raw text, resulting in a representation that contains both structural and linguistic information. This also allows our model to give preference to certain features and relations in specific cases.
The combined encodings are then passed through a \textbf{feed-forward neural network}. The aim here is to collapse the dimension of the model from \textit{\({2} * \text{architecture\_dim}\)} (obtained after concatenation), back to \textit{architecture\_dim}. While doing this, we also hypothesise that the model prunes all the values with low importance after the concatenation, and only keeps the features and relations of high importance for the decoding part.
\subsection{Decoding}
We use the standard Transformer Decoder architecture described in \cite{vaswani2023attentionneed} as the decoding architecture to facilitate the generation of movie summaries from the learned embeddings. The details of implementation of this decoder can be found in the Appendix \ref{appen: implementation-details}.
\section{Results and Findings}
In this section, we provide the detailed results obtained during our experiments with \textbf{DiscoGraMS}.
\subsection{Evaluation Metrics}
To assess the performance of our proposed models in generating summaries, we employ two widely recognized evaluation metrics: \textbf{ROUGE} and \textbf{BERT Scores}. These metrics provide valuable insights into the quality and effectiveness of the generated summaries in comparison to the reference (gold) summaries. More details about the evaluation metrics can be found in Appendix \ref{appen: eval-metrics}

\section{Implementation Details}
\label{appen: implementation-details}
We used a single NVIDIA RTX 6000 with 50 GB VRAM to train and test our model. The VRAM of the GPU was not enough to load models with a higher context size than 4K. 20 Epochs on the train set take 42 hours to complete, while testing on all 20 epochs takes another 4 hours.
The hyper-parameters used while training are as follows:

- Number of Epochs: 20

- Learning Rate: 0.00001

- Architecture Dimension: 4096

- Sentence Encoder (SE) Dimension: 768

- Longformer Encoder (LE) Dimension: 1024

- Dropout in Attention Layer of Encoder: 0.15

- Number of heads in Encoder side Attention: \indent\indent 8

- Dropout in Attention of Encoding Integration: \indent\indent 0.15

- Number of heads in Attention of Encoding \indent \indent Integration: 8

- Decoder Number of Heads: 8

- Decoder Heads: 6

- Internal Dimension of Decoder: 8192

- Max Sequence Length of the Decoder: 2284

\section{Example of a CaD Graphs from the Dataset.}
\begin{figure*}[ht]
    \centering
    \includegraphics[width=1\linewidth]{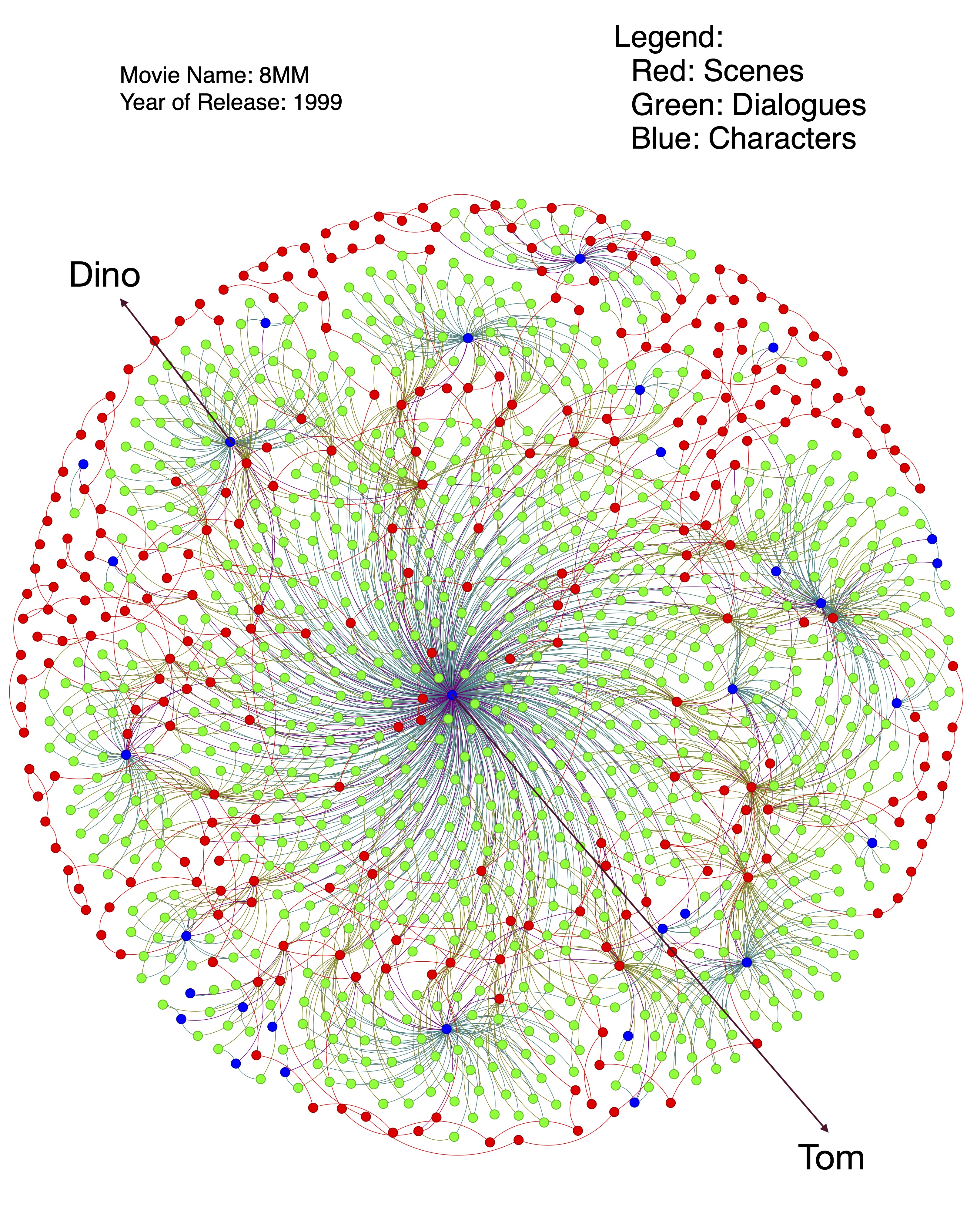}
    \caption{Tom being the Main Protagonist of the movie, naturally has the highest density of edges and is one of the central figures in the graph. This is expected as most of the movie revolves around him. Additionally, a high density can also be observed around the villains such as Dino.}
    \label{fig:8MM}
\end{figure*}

\begin{figure*}[ht]
    \centering
    \includegraphics[width=1\linewidth]{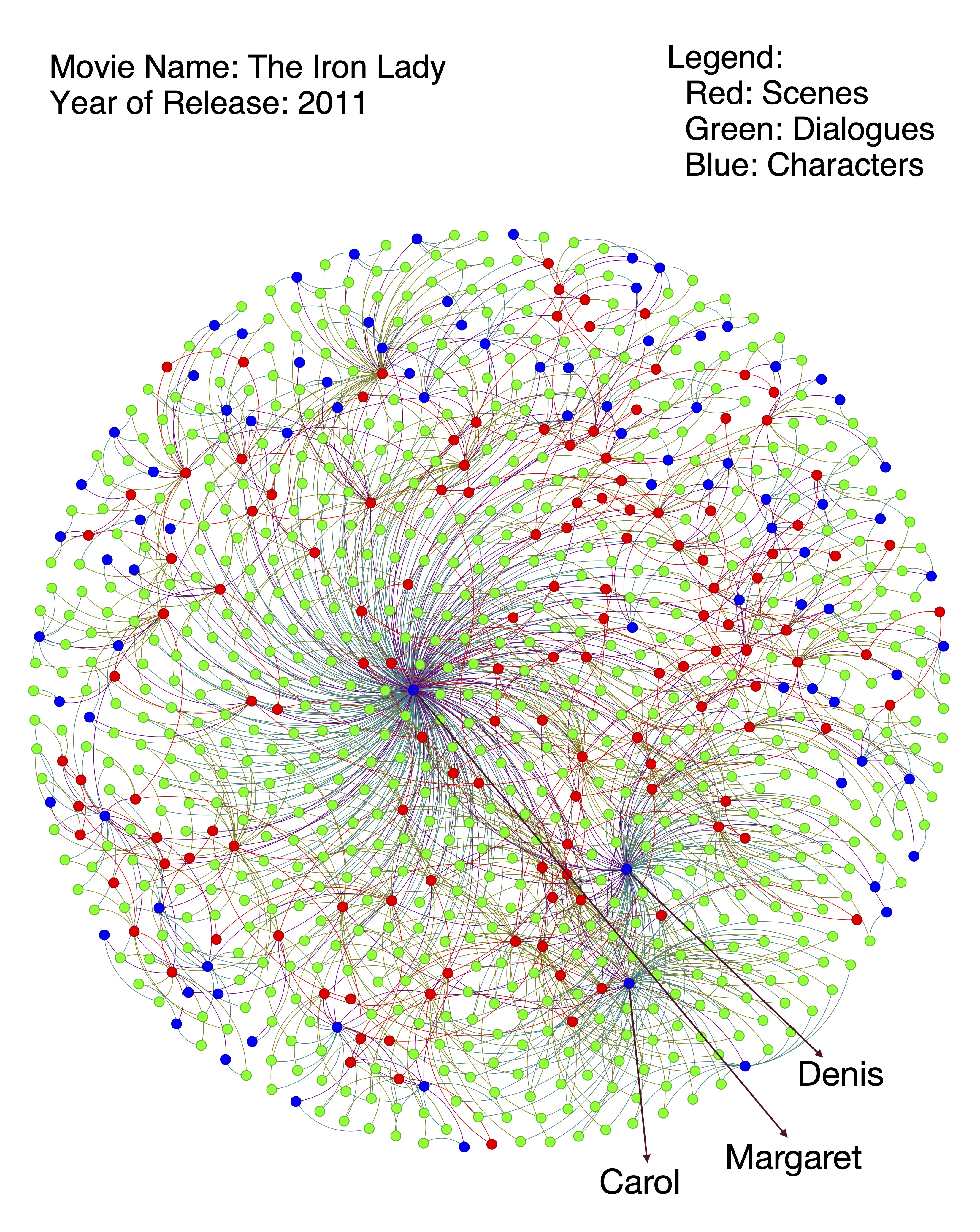}
    \caption{Margaret is the main protagonist of this movie and thus naturally has the highest concentration of edges around her. Additionally, Denis and Carol, her husband and daughter seem to be decently dense as well as they are the immediate family of the main protagonist and they too play an important role in the movie. Owing to the nature of the movie, there is no clear antagonist, and thus, no other major concentration region as well.}
    \label{fig:The Iron Lady}
\end{figure*}

\begin{figure*}[ht]
    \centering
    \includegraphics[width=1\linewidth]{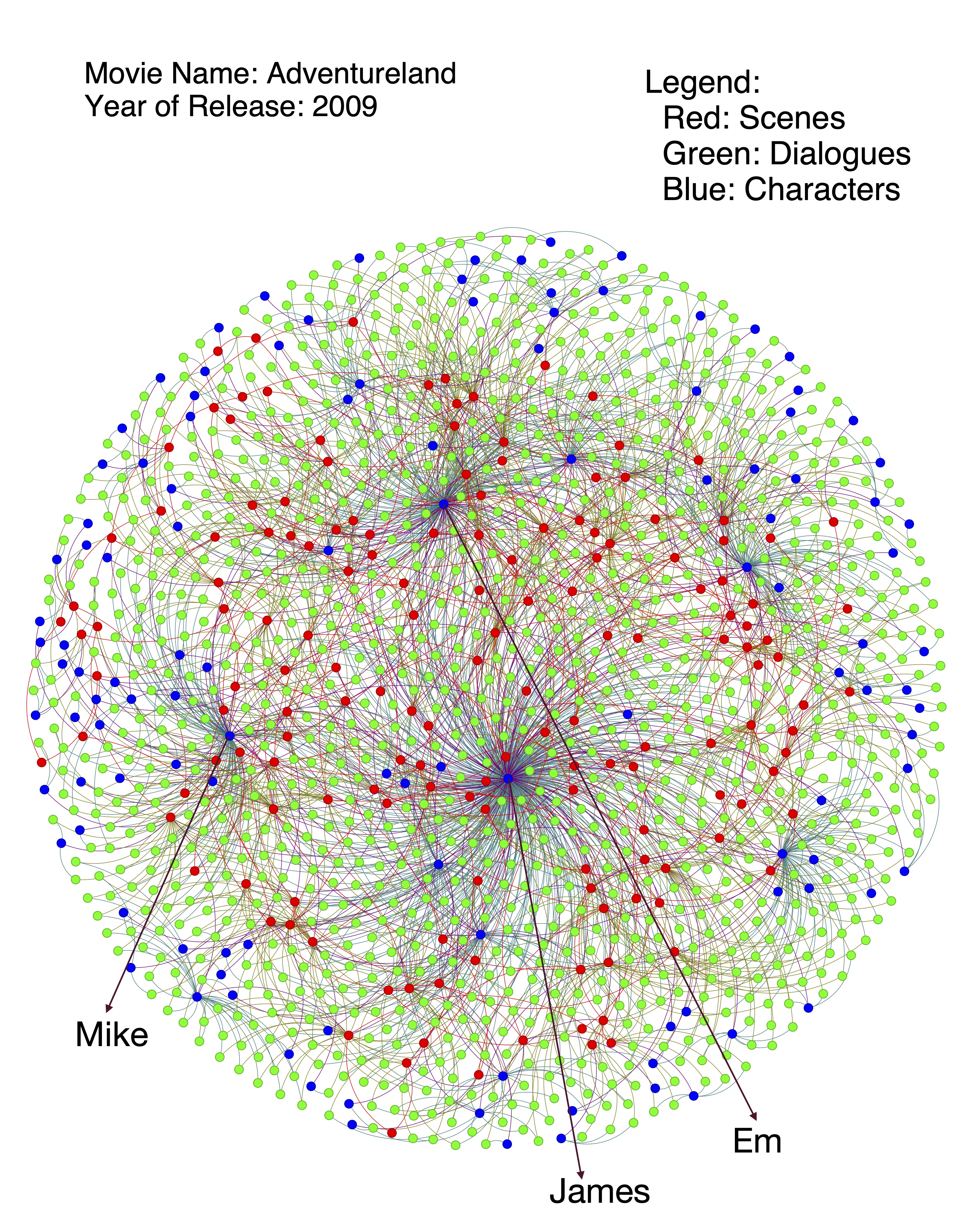}
    \caption{James and Em are the Main Protagonists in the movie, who have a relationship that has bloomed as their summer jobs started at the amusement park Adventureland. Mike is not a traditional villain, but complicates the protagonists relationship as he has an affair with Em. Thus, all three of them have high density edge connections as they contribute to the main density of the movie.}
    \label{fig:Adventureland}
\end{figure*}

In this section, we provide real graphs that we obtain from the dataset used. We visualise these graphs with the help of gephi \footnote{\url{https://gephi.org/}}. Through these examples, we aim to demonstrate our effective character-aware graph construction method and how it helps the model identify the salient characters in the network and the roles that they play. This can be observed by the high density of edges around pivotal characters in the movie. Naturally (or by design), the model will tend to give more importance to these nodes and their connected nodes, deeming them to salient.
\\
- Example graph of the movie \textit{8MM} from 1999 can be seen in Figure \ref{fig:8MM}
\\
- Example graph of the movie \textit{The Iron Lady} from 2011 can be seen in Figure \ref{fig:The Iron Lady}
\\
- Example graph of the movie \textit{Adventureland} from 2009 can be seen in Figure \ref{fig:Adventureland}

\section{Evaluation Metrics}
\label{appen: eval-metrics}
\subsection{ROUGE Scores}
\textbf{ROUGE} (Recall-Oriented Understudy for Gisting Evaluation) Scores \cite{lin-2004-rouge} are a set of metrics used to evaluate automatic summarization and machine translation by comparing the \textbf{overlap of n-grams} between the generated summaries and the reference summaries. We utilize three variants of ROUGE scores:

- \textbf{ROUGE-N}: This measures the overlap of n-grams (where n can be 1, 2, or higher) between the generated summary and the reference summaries. Specifically, ROUGE-1 (Referred to as \textbf{R-1} Later) calculates the overlap of uni-grams, while ROUGE-2 (Referred to as \textbf{R-2} Later) evaluates the overlap of bi-grams.

- \textbf{ROUGE-L}: This metric assesses the longest common sub-sequence between the generated and reference summaries. It captures the fluency of the summary and provides insights into its coherence by considering the order of the words. (This is Referred to as \textbf{R-L} Later)

Higher ROUGE scores indicate better alignment with the reference summaries.

\subsection{BERT Scores}
\textbf{BERT Scores} \cite{bert-score} leverage contextual embeddings derived from the BERT model \cite{devlin2019bertpretrainingdeepbidirectional} to evaluate the quality of generated summaries. Unlike traditional n-gram-based methods, BERT scores take into account the \textbf{semantic similarity} between the generated and reference summaries. BERT Scores are usually reported as:

- BERT Score Precision (\textbf{BS$_p$}): It focuses on the accuracy of the generated content.

- BERT Score Recall (\textbf{BS$_r$}): It emphasizes completeness in capturing relevant content.

- BERT Score F1 Score (\textbf{BS$_{f1}$}): It combines both metrics to provide a balanced assessment of summary quality
\\
\\
By utilizing both ROUGE and BERT scores, we can gain a well-rounded understanding of how our proposed models perform in terms of both surface-level text overlap and deeper semantic alignment with gold summaries. This dual approach allows for a more robust evaluation of the generated summaries, ensuring that they not only contain relevant information but also maintain coherence and fluency.

\end{document}